\documentclass[pdflatex,sn-mathphys-num]{sn-jnl}%

\usepackage{graphicx}%
\usepackage{multirow}%
\usepackage{amsmath,amssymb,amsfonts}%
\usepackage{amsthm}%
\usepackage{gensymb}
\usepackage{siunitx}
\usepackage{mathrsfs}%
\usepackage[title]{appendix}%
\usepackage{xcolor}%
\usepackage{textcomp}%
\usepackage{manyfoot}%
\usepackage{booktabs}%
\usepackage{algorithm}%
\usepackage{algorithmicx}%
\usepackage{algpseudocode}%
\usepackage{listings}%
\usepackage{natbib}
\usepackage{hyperref}
\usepackage[capitalize]{cleveref}
\usepackage[normalem]{ulem}
\usepackage[font=small, labelfont=small]{caption}
\useunder{\uline}{\ul}{}

\begin{document}

\title{All Eyes, no IMU: Learning Flight Attitude from Vision Alone}

\author[1]{\fnm{Jesse J.} \sur{Hagenaars}}
\equalcont{Equal contribution; author ordering determined by coin flip.}

\author*[1]{\fnm{Stein} \sur{Stroobants}}\email{s.stroobants@tudelft.nl}
\equalcont{Equal contribution; author ordering determined by coin flip.}

\author[2]{\fnm{Sander M.} \sur{Boht\'e}}

\author[1]{\fnm{Guido C.H.E.} \sur{de Croon}}

\affil[1]{\orgdiv{MAVLab}, \orgname{TU Delft}}
\affil[2]{\orgdiv{ML group}, \orgname{CWI}}

\abstract{
Vision is an essential part of attitude control for many flying animals, some of which have no dedicated sense of gravity. Flying robots, on the other hand, typically depend heavily on accelerometers and gyroscopes for attitude stabilization. In this work, we present the first vision-only approach to flight control for use in generic environments. We show that a quadrotor drone equipped with a downward-facing event camera can estimate its attitude and rotation rate from just the event stream, enabling flight control without inertial sensors. Our approach uses a small recurrent convolutional neural network trained through supervised learning. Real-world flight tests demonstrate that our combination of event camera and low-latency neural network is capable of replacing the inertial measurement unit in a traditional flight control loop. Furthermore, we investigate the network's generalization across different environments, and the impact of memory and different fields of view. While networks with memory and access to horizon-like visual cues achieve best performance, variants with a narrower field of view achieve better relative generalization. Our work showcases vision-only flight control as a promising candidate for enabling autonomous, insect-scale flying robots.
}

\maketitle

\section{Introduction}
\label{sec:introduction}

\begin{figure}[!t]
    \centering
    \includegraphics[width=\linewidth]{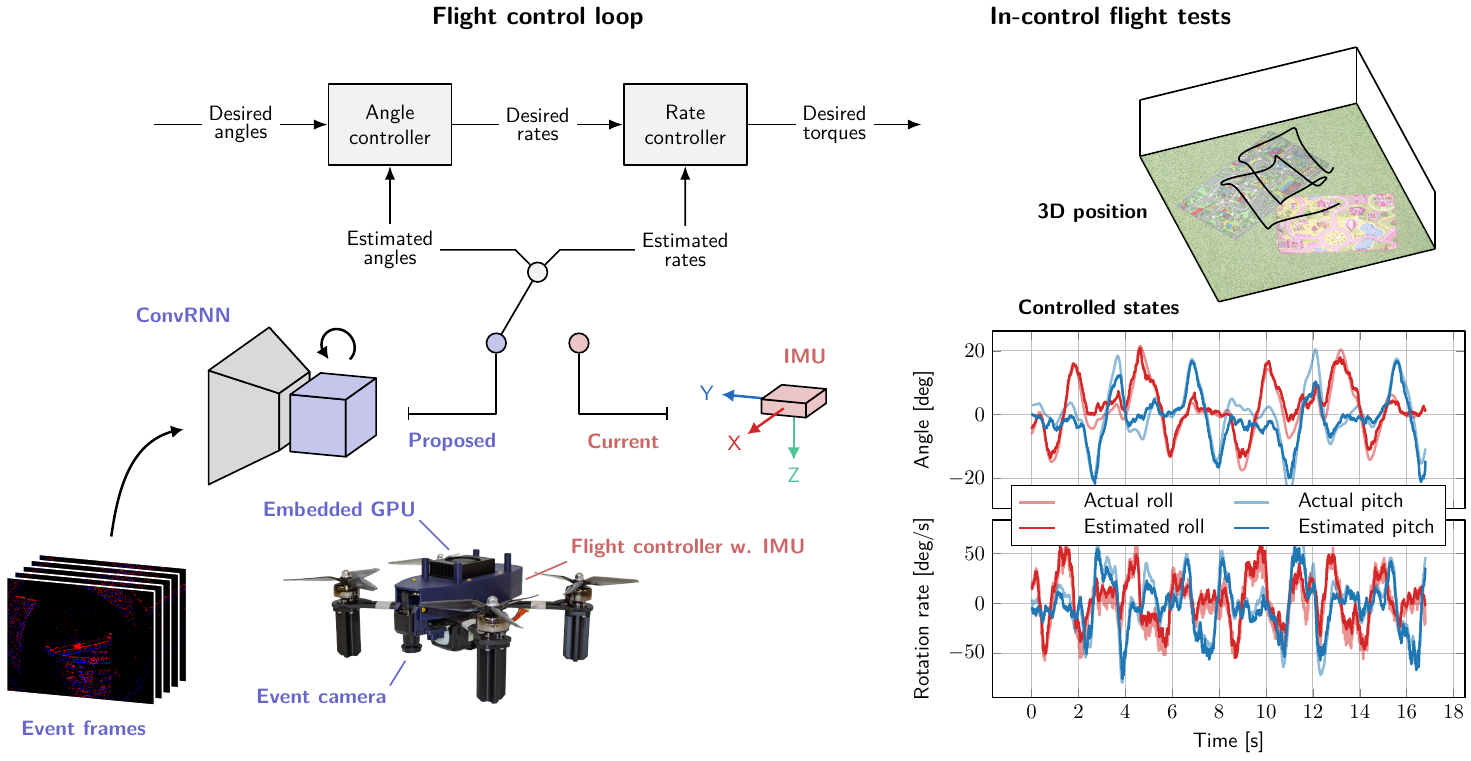}
    \caption{Vision-only flight control promises lighter flying systems. We demonstrate that fully on-board, vision-only flight control is possible with a low-latency vision pipeline consisting of a small recurrent convolutional neural network (ConvRNN) and an event-based camera. Left: Our proposed pipeline takes the place of the IMU (inertial measurement unit) in a traditional flight control loop, estimating both attitude and rotation rates. Right: The resulting system demonstrates accurate estimation and control in real-world flight tests. The plots show both the actual attitude angles and rates (as measured by the flight controller) and the estimated ones (predicted by a neural network).}
    \label{fig:overview}
\end{figure}

Attitude control is a fundamental challenge in aerial robotics. For drones to execute their missions, they must precisely control their orientation relative to gravity---a task traditionally addressed by IMUs (inertial measurement units) that provide absolute acceleration and rotation rate measurements~\cite{mahony2008nonlinear}. Yet, flying insects exhibit remarkable flight agility without any known sensor dedicated to measuring gravity~\cite{taylor2007sensory}. Flying insects with four wings such as honeybees even lack the halteres that provide two-winged flying insects with direct sensory information on rotation rates~\cite{srinivasan2012biology}. Previous work has hypothesized that flying insects can in principle rely on visual cues alone to estimate flight attitude~\cite{decroon2022accommodating}. Specifically, it was shown that optical flow can be combined with a motion model to estimate and control flight attitude. Besides insect understanding, this insight opens a pathway toward lighter and potentially more robust flying robots. By eliminating the dependency on IMUs, ultra-lightweight sensor suites~\cite{yu2025tinysense} could be made even lighter and more efficient. This work demonstrates a general approach to flight control without IMU---bringing tiny autopilots and insect-scale flying robots one step closer.

Vision-based attitude estimation for flying robots goes back to early work on horizon-line detection methods, applied to fixed-wing drones flying high in wide open environments~\cite{ettinger2003visionguided, mondragon2010unmanned}. 
Later, methods have been developed that rely on the specific structure of human-made environments. Assuming parallel lines in view, vanishing points can be determined and used as attitude estimators~\cite{bazin2008uav, shabayek2012vision}. 
However, flying insects are also able to control their attitude in unstructured environments where the sky is not visible, a property that is also of interest for flying robots. 
Combining optical flow with a motion model enables attitude estimation in such generic environments, only relying on sufficient visual texture. 
De Croon et al.~\cite{decroon2022accommodating} show that attitude can be inferred from optical flow when combined with a motion model that relates attitude to the direction of acceleration. However, due to hardware-related update-rate limitations, their real-world flight demonstrations still rely on gyroscope measurements.

Event-based cameras offer a promising solution to these limitations~\cite{gallego2020eventbased} with their low latency, high temporal resolution, and robustness to motion blur. Existing work on estimating rotation rates with event cameras~\cite{gallego2017accurate,gehrig2020eventbased,greatorex2025eventbased} has so far been limited to motions with little translation-rotation ambiguity (rotation-only or restricted motion like driving). Furthermore, work on estimating flight attitude from vision has been limited in environmental complexity (requiring a structured environment with vanishing lines)~\cite{dacosta2025gyrevento} or still requires the use of gyroscopes for low-level flight control~\cite{geles2024demonstrating,xing2024bootstrapping,romero2025dream}.
Thus, a critical gap remains: achieving fully on-board, IMU-free attitude control in realistic flight conditions.

In this paper, we address this gap by demonstrating---for the first time---a fully on-board flight control system capable of estimating both attitude and rotation rate solely from event-based visual inputs.
Specifically, we develop a recurrent convolutional neural network trained through supervised learning to map raw event streams directly to accurate attitude and rotation rate estimates. 
Unlike traditional approaches, which use IMU measurements in combination with a filter to explicitly model the drone's motion, we take a learning approach which trains a neural network to implicitly acquire the relation between visual cues and the vehicle's state. Opting for learning a neural network means our estimator can eventually be integrated in an autopilot which is learned end-to-end as a neural network.
This will be especially relevant if not only event-based vision but also neuromorphic computing is used on the flying robot~\cite{paredes-valles2024fully,stroobants2025neuromorphic}.

We demonstrate, with real-world flight, that our combination of event camera and low-latency neural network can replace the traditional IMU and filter combination in the flight control loop. Furthermore, we investigate the learned network's generalization to different environments, and compare alternative network inputs and architectures. We show that neural network memory and a wide field of view are essential components for accurate estimation, but that greatly reducing the field of view, and denying the network of most horizon-like visual cues, leads to improved relative generalization across environments. While this hints at the learning of an internal model for attitude from visual motion, it comes at the cost of reduced absolute performance.

The contributions of this work can be summarized as follows:
\begin{enumerate}
\item The first fully on-board, vision-only and IMU-free pipeline for control of a real-world, unstable quadrotor.
\item A low-latency recurrent convolutional neural network capable of estimating flight attitude and rotation rate from event camera data through supervised learning.
\item An investigation into the approach's performance, necessary components, and generalization to different environments.
\end{enumerate}

Our work promises extremely efficient, end-to-end-learned autopilots with minimal sensors, capable of powering the next generation of insect-scale flying robots.

\section{Results}

\subsection*{In-the-loop flight tests}

\cref{fig:overview} gives an overview of our proposed system. We integrate an event camera and small recurrent convolutional neural network running on an on-board GPU into the drone's flight control loop. The network estimates attitude and rotation rates from event camera data with low latency, acting as a stand-in replacement of a regular IMU (inertial measurement unit), and allowing for accurate control. In traditional flight control loops, IMUs measure linear accelerations and rotation rates at high frequency (typically $\gg$ 500~Hz). 
These measurements are subsequently integrated in a Kalman or complementary filter running at a lower frequency (100-200 Hz) to produce accurate attitude and rotation rate estimates while filtering out the high-frequency noise produced by the sensors.

The proposed system instead uses data from an event camera in combination with a learned estimator. Events are accumulated into frames of 5~ms, and fed to a recurrent convolutional neural network that then estimates attitude and rotation rate. These estimates are sent to the flight controller at 200~Hz, and used by the regular angle and rate controller to control the drone. We train our network through supervised learning on a dataset containing events, along with attitude and rotation rates from the flight controller as labels. While these are not absolute ground truth (they are estimated by the flight controller using the IMU), they are typically reliable but can have bias (which is dealt with by subsequent integrators).

\begin{figure}[t]
    \centering
    \includegraphics[width=\linewidth]{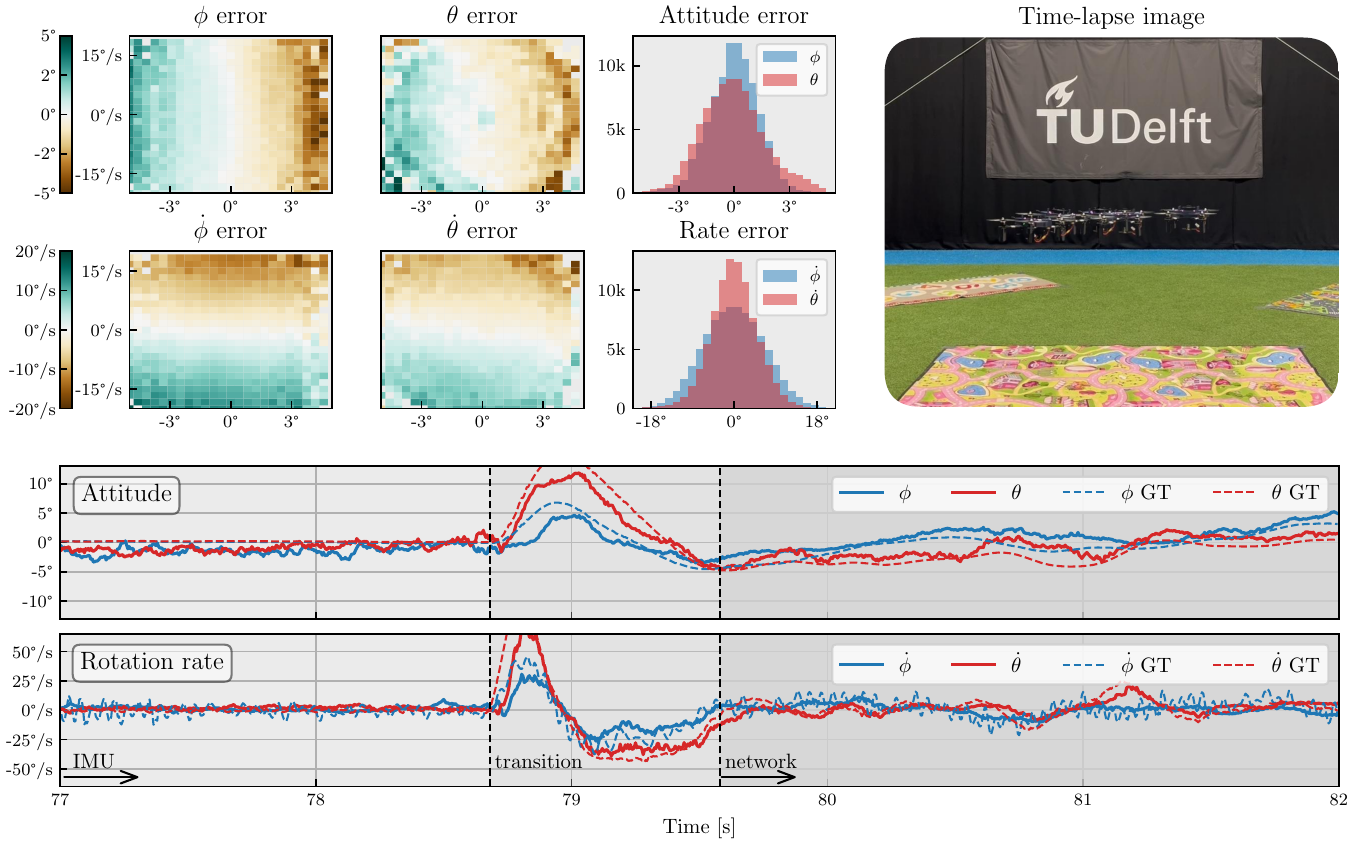}
    \caption{In-the-loop hover (position hold) tests with the network in control. The right-most time-lapse image shows the variation of the drone during flight, demonstrating the controllability of the drone over a total of approx. 10 minutes of flight time. The other plots quantify the errors of the estimated attitude and rotation rate for roll (rate) $\phi$ ($\dot{\phi}$) and pitch (rate) $\theta$ ($\dot{\theta}$). We consider the flight controller estimator as ground truth (GT). The histograms give the error distribution of the network estimate compared to ground truth, with the biases of both subtracted (to disregard biases in training data). The left-most plots show the network errors for different attitude/rate combinations. These show that the network for both attitude and rate underestimates the real value. The bottom traces show the transition from IMU to network control, with a reset of the flight controller's integrator causing a short response of the drone before settling.}
    \label{fig:network_hover}
\end{figure}

The results from several flight tests with the trained network in the loop are shown in \cref{fig:network_hover}. For these tests, the pilot commands the drone to hold its position, which the flight controller translates to attitude setpoints with the help of a higher-level position controller and a velocity sensor. These attitude setpoints are subsequently compared against estimates provided by the neural network. This results in desired rotation rates, which are also compared against network estimates, resulting in desired torques sent to the motors. \cref{fig:network_hover} contains flight tests in which the drone is commanded to stay in the same place. 

The traces and histograms show that the network can accurately estimate both attitude and rotation rate, with most errors within $\pm3$~deg and $\pm18$~deg/s for approx. 10 minutes of flight. The error plots for different attitude/rate combinations show that most errors are due to underestimation by the network, which can also be seen in the difference between network and ground truth during the transition phase. Underestimation is most pronounced at high angles/rotation rates, which could be explained by the inertia of the network's memory. While this allows integration of information over time, it also limits the network in following fast maneuvers.

The error histograms further reveal axis-dependent asymmetries: attitude estimates are more accurate for roll than for pitch, whereas the opposite holds for rotation rate estimates. We attribute reduced accuracy in pitch attitude estimation primarily to limited pitch-angle variability during training and minor shifts in drone balance along the pitch axis (such as varying battery position).
Conversely, the decreased accuracy observed in roll rate predictions aligns with the higher noise levels present in the ground-truth roll rate measurements coming from the flight controller.

\subsection*{Comparison of models}

\begin{figure}[htbp]
  \centering
  \begin{minipage}[c]{0.29\textwidth}
    \centering
    \includegraphics[width=\linewidth]{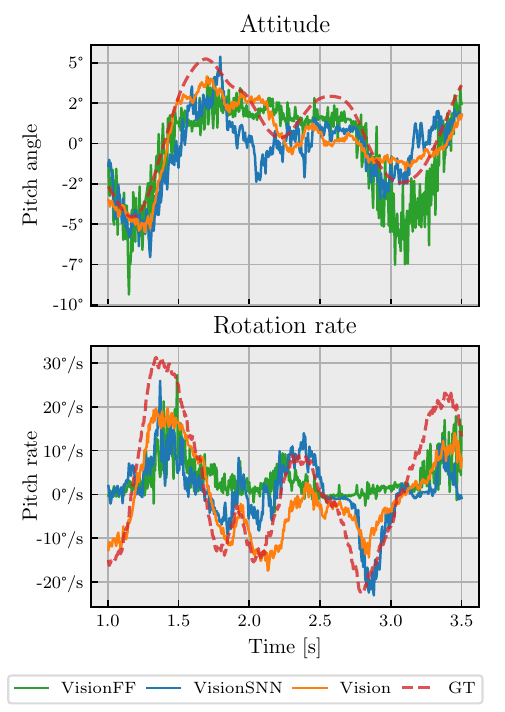}
    \caption{Qualitative comparison of attitude and rotation rate estimates for various neural network architectures on unseen data from the training environment. The network variants match those in \cref{tab:results}.}
    \label{fig:network_traces}
  \end{minipage}
  \hfill
  \begin{minipage}[c]{0.69\textwidth}
    \centering
    \captionof{table}{Performance of models with different network architectures and inputs on unseen data from the training environment. \textit{Vision} is the baseline model with ConvGRU memory and vision-only input. \textit{VisionMotor} additionally has motor commands as input. \textit{VisionGyro} receives gyro measurements as additional input. \textit{VisionFF} has a memory-less architecture (feedforward). \textit{VisionSNN} is a hybrid spiking neural network. We quantify performance in terms of RMSE (root mean square error) and MASD (mean absolute successive difference).}
    \label{tab:results}
    \begin{tabular}{lrrrr}
      \toprule
                  & \multicolumn{2}{c}{Attitude}    & \multicolumn{2}{c}{Rotation rate} \\ 
      \cmidrule(l){2-5} 
      Network     & RMSE {[}deg{]} & MASD {[}deg{]}  & RMSE {[}deg/s{]}  & MASD {[}deg/s{]}   \\ 
      \midrule
      Vision      & {\ul 1.51}    & \textbf{0.27} & 10.65          & 2.57           \\
      VisionMotor & 1.64          & {\ul 0.31}          & {\ul 9.57}     & {\ul 2.47}     \\
      VisionGyro  & \textbf{1.47} & {\ul 0.31}    & \textbf{4.01}  & \textbf{2.36}  \\
      VisionFF    & 2.17          & 1.00          & 18.65          & 5.82           \\
      VisionSNN   & 2.17          & 0.52          & 15.03          & 4.04           \\ 
      \bottomrule
    \end{tabular}
  \end{minipage}
\end{figure}

We conducted an extensive comparison of neural network models with varying network architecture, input modalities and input resolution. \cref{tab:results} lists the performance of these variations in terms of RMSE (root mean square error) and MASD (mean absolute successive difference) when tested on unseen data from the training environment. While RMSE gives a good measure of the estimation error, MASD quantifies prediction smoothness by looking at the difference between subsequent predictions. A qualitative illustration of the estimation performance of various neural networks is given in \cref{fig:network_traces}.

The baseline, \textit{Vision}, receives only event frames as input and processes these with a convolutional neural network with GRU (gated recurrent unit) memory block. This variant was used for the in-control flight tests in \cref{fig:network_hover}. \textit{VisionMotor} additionally takes motor speeds as input. These can serve as a proxy for the moment generated by the motors, (a prediction of) which was shown to be necessary for the attitude to be observable~\cite{decroon2022accommodating}. In our learning setup, however, the error difference with the vision-only model is small: only the estimation of rotation rates is slightly better, which makes sense given that the forces produced by the motors can be integrated to obtain rotational velocities. \textit{VisionGyro} receives gyro measurements (rotation rates) as additional inputs. This represents the case in which a gyro would still be present in the system, and its performance can give an idea of whether the network could integrate rotation rate to obtain the attitude. As expected, this results in the lowest-error rotation rate estimates. The accuracy of the attitude prediction, however, is not meaningfully better.
\textit{VisionFF} replaces the recurrent memory block with a feedforward alternative, and will therefore not be able to integrate information over time. While attitude can be inferred from a single image by looking at horizon-like visual cues (such as those indicated by the white arrows in \cref{fig:motionmodel}), the increased attitude error of \textit{VisionFF} compared to \textit{Vision} suggests that having memory is still beneficial. Estimation of velocities such as rotation rate is very difficult without memory, and this is reflected by the large increase in rotation rate error. 

\textit{VisionSNN} is a hybrid spiking neural network (SNN) where the encoder has binary activations (stateless spiking neurons) and the recurrent memory block consists of spiking LIF (leaky integrate-and-fire) neurons with a recurrent connection. While the quantitative errors for \textit{VisionFF} and \textit{VisionSNN} are similar, \cref{fig:network_traces} shows that the SNN's memory makes a difference for estimating rotation rates.

\begin{figure}[h!]
    \centering
    \includegraphics[width=\linewidth]{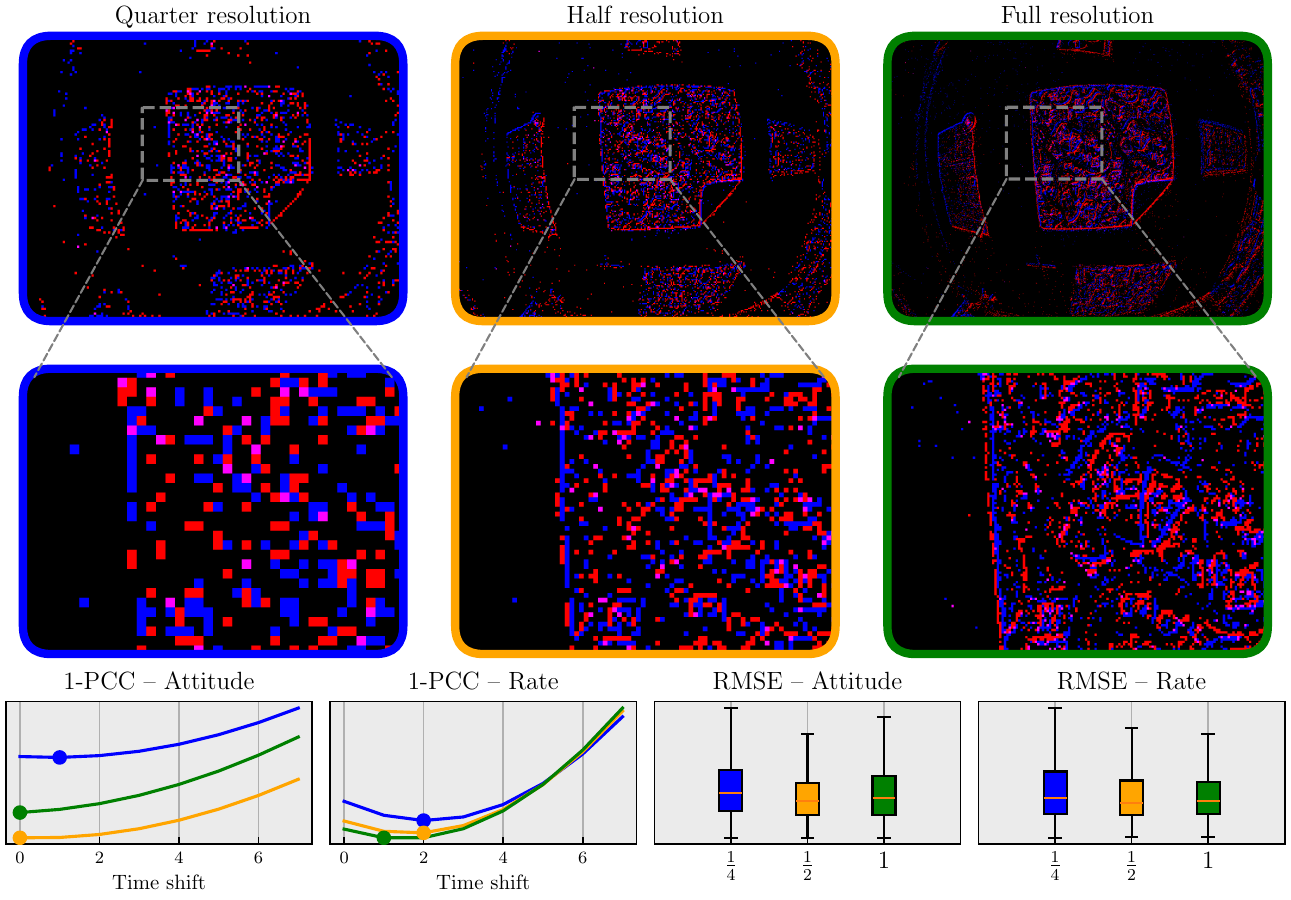}
    \caption{The impact of resolution on network performance after training under otherwise identical settings. Results were obtained by enabling only a subset of event camera pixels: every fourth pixel (quarter resolution, blue), every other pixel (half resolution, orange) and all pixels (full resolution, green). The bottom-left graphs show the PCC (Pearson correlation coefficient) for different time shifts of the prediction targets. The minimum of each line indicates the shift with the highest correlation. Minima at larger shifts indicate a delay in the network prediction. The bottom-right plots show the RMSE (root mean square error) on validation data. Networks trained on quarter-resolution data show larger prediction delay and increased error compared to higher resolutions.}
    \label{fig:resolution_comp}
\end{figure}

Next, we investigate the impact of varying input resolution. Under rapid motion, event cameras generate a large amount of events, which can lead to bandwidth saturation and increased latency when working with on-board, constrained hardware. The event camera used in this work, a DVXplorer Micro, allows disabling pixels to limit the number of events generated by the camera. \cref{fig:resolution_comp} analyzes the impact of using lower-resolution data on network performance when training with otherwise identical settings. Apart from quantifying the estimation error using RMSE, we also look at the prediction delay of the network. When actively controlling a system, significant delays lead to oscillations and potentially instability, and should therefore be avoided. We quantify the prediction delay of each network by looking at the time shift for which the Pearson correlation coefficient (PCC) is lowest. The results show that there is a noticeable delay in predictions when using only a quarter of the camera's pixels. We attribute this to the fact that attitude changes might only be seen by any of the enabled pixels once they become large enough, leading to a delayed response. This delay is much less present for half and full-resolution networks. Interestingly, the full-resolution network shows a slightly higher attitude RMSE compared to the half-resolution network. This could be explained by the fact that all networks were trained for the same number of epochs, whereas the larger full-resolution network likely requires more training steps before achieving similar convergence. The half-resolution network brings a good balance between computational efficiency and estimation performance.

\subsection*{Generalization and internal motion model}

\begin{figure}[h!]
    \centering
    \includegraphics[width=\linewidth]{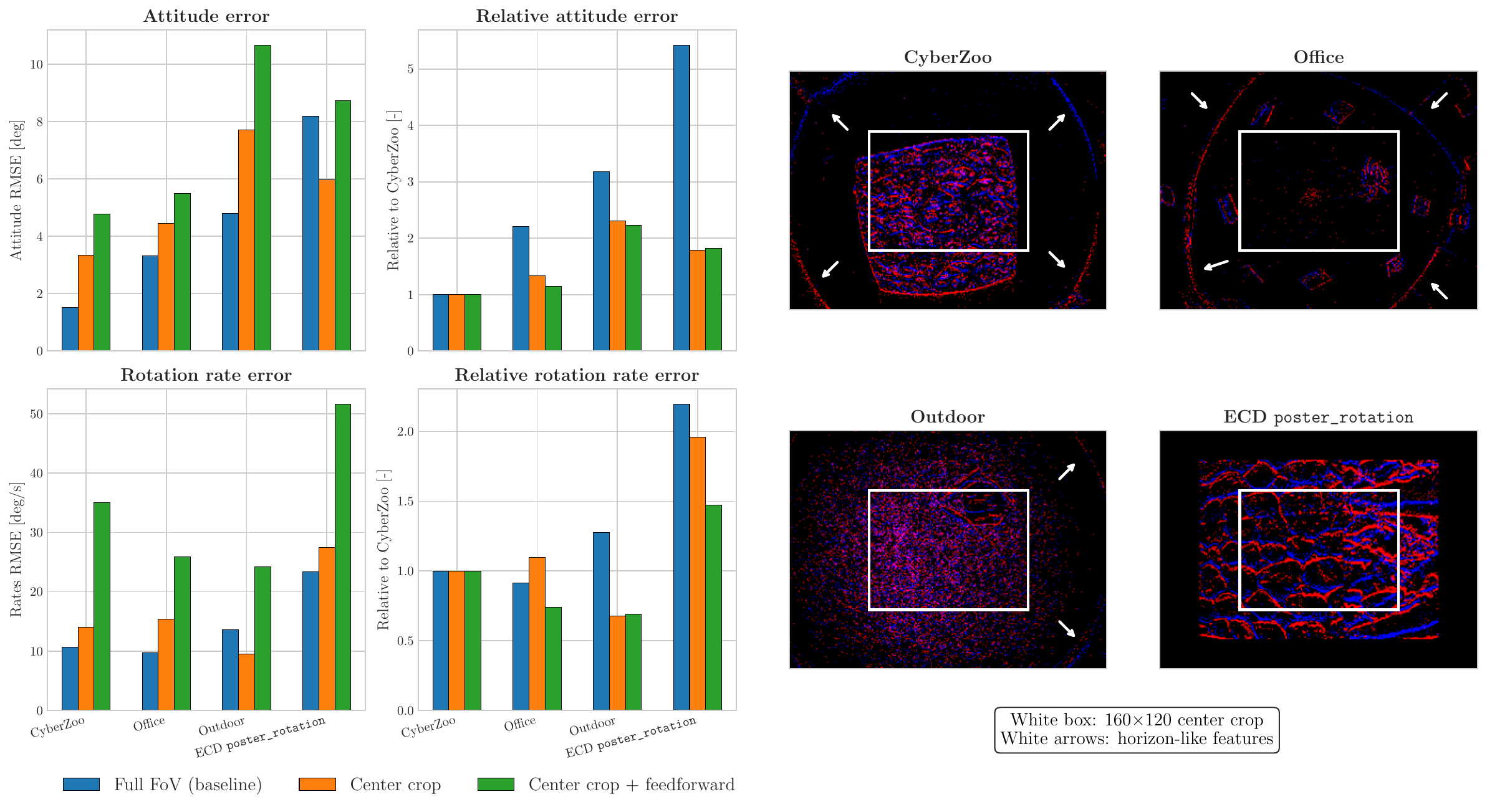}
    \caption{Comparison between a network with memory and full field-of-view (baseline), a network with memory that only sees a center-cropped portion (indicated by the white rectangle), and a network without memory that only sees a center-cropped portion. We compare estimation errors for four unseen sequences. CyberZoo is the same environment as trained on, but Office, Outdoor and ECD \texttt{poster\_rotation} (rotation only)~\cite{mueggler2017eventcamera} are unseen environments. While the full-FoV network performs better in scenes where horizon information (indicated by white arrows) is available to provide an absolute indication of attitude (CyberZoo, Office, Outdoor), the center-cropped network performs better when there are no visible horizon-like cues (as in ECD \texttt{poster\_rotation}), hinting at the use of an internal motion model. Additionally, the smaller relative increase in error in the case of the center-cropped network for scenes different from the training location CyberZoo indicates improved generalization. A memory-less center-crop network is unable to aggregate information internally into any kind of model, and hence performs poorly in terms of both attitude and rotation rate estimation.}
    \label{fig:motionmodel}
\end{figure}

Recent work~\cite{decroon2022accommodating} has shown that attitude can be inferred from optical flow when combined with a motion model relating attitude to acceleration direction. While the learning framework presented here does not explicitly represent such a model internally, it would be interesting to investigate whether this can be promoted during learning, and whether this affects generalization to different scenes. In other words, we would like to steer the network towards using optical flow for attitude estimation, instead of static visual cues such as horizon-like straight lines on the edges of the field of view (indicated by white arrows on the right in \cref{fig:motionmodel}). We hypothesize that networks that rely mainly on motion features generalize better across environments than networks that focus on visual appearance, which is more scene-specific and may lead to overfitting on the training scene.

To achieve this, we trained a network on the same dataset as before (CyberZoo), but restricted its input to a small 160$\times$120 center crop. This eliminates most visual cues that contain absolute attitude information while preserving motion cues. We refer to the original, unrestricted model as \textit{full FoV} (identical to \textit{Vision} from \cref{tab:results}) and the newly trained variant as \textit{center crop}. We evaluate these models on four unseen sequences, and show the results in \cref{fig:motionmodel}. The full-FoV network effectively makes use of the horizon-like cues present in most sequences (white arrows for CyberZoo, Office and Outdoor), generalizing well to other scenes and outperforming the center-crop model. However, we also include a sequence from the event camera dataset ECD \texttt{poster\_rotation} recording~\cite{mueggler2017eventcamera}. Here, a camera (different from ours) looks at a planar poster while undergoing rotations, without any visual cues related to the camera's attitude. On this sequence, the center-crop network achieved lower attitude and rate errors than the full-FoV network. This indicates that the center-crop network can effectively use motion information to infer attitude, hinting at an internally acquired motion model. Furthermore, looking at the relative error across environments, we see that relative error increases less for the center-crop network when moving from the familiar training environment to novel scenes. This presents a trade-off: while networks with access to the full FoV have better absolute performance for scenes with horizon-like visual cues, networks forced to infer attitude from motion through a reduced FoV relatively generalize better across environments. If we remove memory from the network (\textit{center crop + feedforward}), it performs poorly for both attitude and rotation rate estimation. Such a network has neither the field of view for static attitude information, nor the ability to build it up through memory.

\section{Discussion and conclusion}

In this work, we presented the first demonstration of stable low-level flight control based purely on visual input, eliminating the need for an IMU (inertial measurement unit). By combining an event camera with a compact recurrent convolutional neural network, we achieved real-time attitude and rotation rate estimation, enabling closed-loop control without inertial sensing. Leveraging the event camera's low latency and high temporal resolution, our vision pipeline delivers the responsiveness required for agile flight, running entirely on-board at 200~Hz.

Removing the IMU simplifies hardware, reducing weight and energy consumption---both critical considerations for small, bio-inspired flying robots---and our work shows that it is possible to estimate and control flight attitude based on vision alone. Such a vision-only control pipeline offers key advantages for aerial robotics: insect-scale flying robots will already rely on visual inputs for navigation, so with our proposed method no extra sensors would be necessary. Processing could run on a tiny energy-efficient, possibly neuromorphic, processor. Our experiments show that estimation performance generalizes to unseen and visually distinct environments, and that this can potentially be improved further by forcing the network to infer attitude from visual motion instead of horizon-like appearance cues. Comparisons between network architectures further highlight the importance of memory: while feedforward networks can infer static attitude from scene appearance, recurrent models are crucial for accurately tracking dynamic flight states, underlining the role of memory in enabling robust vision-based control.

Several limitations remain. We observed systematic underestimation of both attitude and rotation rates, with axis-dependent asymmetries. These can be attributed to biases in training data distribution, particularly in pitch motions, and to shifts in drone balance. Reducing these biases---through more diverse datasets, improved calibration or higher-accuracy ground-truth (such as from a motion capture system)---could improve overall performance. Additionally, we found that lowering input resolution introduces prediction delays, while higher resolutions impose greater computational demands without proportional gains in performance. A half-resolution setting emerged as a practical trade-off for real-time operation.

Future work should focus on increasing the robustness of learning and further hardware integration. Making use of the contrast maximization framework for self-supervised learning from events~\cite{gallego2018unifying} would allow direct learning of a robust and low-latency estimator of optical flow decomposed as rotation and translation~\cite{paredes-valles2024fully}. This could be integrated with a learned visual attitude estimator to give the most robust estimate, implicitly combining both visual motion as well as horizon-like features. Hardware could further be integrated through a combined event camera and neuromorphic processor setup, such as the SynSense Speck~\cite{richter2024speck}, to achieve lower power consumption, weight and latency.
Furthermore, a deeper investigation into the internal motion model developed by networks may yield deeper insights into the mechanisms they use to solve the task at hand. We have shown that parts of the network generalize across scenes, but also that the networks can exploit scene-specific features that resemble horizon-like lines to get an estimate of the attitude. Such insights could enable the design of more robust and autonomous robotic systems that rely heavily on environmental perception.

Overall, our results demonstrate that vision-only attitude estimation and control is a viable alternative to traditional inertial sensing. By simplifying the sensor suite and removing the dependency on IMUs, our approach paves the way for the next generation of lightweight, agile, and bio-inspired flying robots.

\section{Methods}
\label{sec:methods}

\subsection*{Estimating attitude and rotation rate from events}

The analysis by De Croon et al.~\cite{decroon2022accommodating} shows that a drone's flight attitude can be extracted from optical flow when combined with a motion model. Many conditions are analyzed. In the simplest case, they derive local observability from the ventral optical flow component $\omega_y$ for roll $\phi$ and a simple motion model that relates attitude angles to acceleration direction:
\begin{align}
    \omega_y &= - \frac{\cos^2(\phi)\upsilon_y }{z} + p \label{eq:g1} \\
    f(\boldsymbol{x}, u) &=
    \begin{bmatrix}
        \dot{\upsilon}_y \\
        \dot{\phi} \\
        \dot{z}
    \end{bmatrix} = 
    \begin{bmatrix}
        g\tan(\phi) \\
        p \\
        0
    \end{bmatrix} \label{eq:g2}
\end{align}

where state $\boldsymbol{x} = [\upsilon_y,\phi,z]^T$, control input $u=p$, $z$ represents the height above ground, $\upsilon_y$ is the velocity in the body frame's $y$-direction, and $p$ denotes the roll rate. This relationship holds for $\phi \in (\ang{-90},\ang{90})$, and the same derivation can be made for $\omega_x$ and pitch $\theta$. Although this model assumes constant height, they show extensions for changing height and uneven/sloped environments. In those, the divergence of the optical flow field can be used to observe variation in height, maintaining the partial observability of the system. The system is only partially observable, since attitude becomes unobservable at angles and rates close to zero, for instance when the drone is hovering in place. Nevertheless, the parts of the state space that make the system unobservable are inherently unstable and drive the system to observable states, thus closing the loop. 

While \cref{eq:g1,eq:g2} treat the rotation rate $p$ as a known control input (from gyro measurements), this is theoretically not necessary, and a prediction of the moments $M$ resulting from control inputs suffices~\cite{decroon2022accommodating}:
\begin{equation}
    f(\boldsymbol{x}, u) =
    \begin{bmatrix}
        \dot{\upsilon}_y \\
        \dot{\phi} \\
        \dot{p} \\
        \dot{z}
    \end{bmatrix} = 
    \begin{bmatrix}
        g\tan(\phi) \\
        p \\
        M/I \\
        0
    \end{bmatrix}
\end{equation}

where $I$ is the moment of inertia around the relevant axis. In our approach, motor speeds as input could replace $M$ and the network could learn an internal motion model to obtain attitude. However, this is not guaranteed, and as the successful flight tests with a vision-only network show, not necessary either when using a machine learning approach.
Learned models exploit other factors, as may insects. We explore one of these factors (a large field of view) in this article.
A large field of view enables proper separation of translational and rotational flow, which allows more accurate extraction of control inputs from vision. As described above, these control inputs allow for the system to be observable.

\subsubsection*{Training and network details}

\begin{figure}[t]
    \centering
    \includegraphics[width=\linewidth]{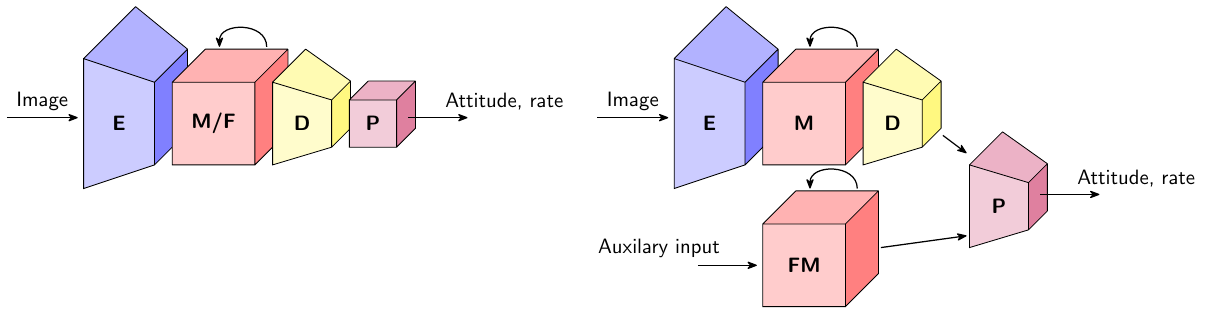}
    \caption{Schematic overview of network architectures. Left: baseline vision-only network. Right: network with vision and an auxiliary input (motor speeds, rotation rates). Encoder (E), memory (M) and decoder (D) are convolutional. Flattening to attitude and rotation rate estimates happens in the predictor (P). We swap the memory for a feedforward (F) block to get a network without recurrency. For the auxiliary input, we use fully connected (FM) instead of convolutional memory.}
    \label{fig:networks}
\end{figure}

We train a small recurrent convolutional network to estimate flight attitude and rotation rate from events. Training is done in a supervised manner, with ground-truth attitude and rate coming from the flight controller's state estimation. We collect training data containing diverse motions and attitudes in our indoor flight arena. We make use of two-channel event count frames as input to the network, with each frame containing 5~ms of events. For training, we take random slices of 100 frames from a sequence, and use truncated backpropagation through time on windows of 10 frames without resetting the network's memory. This is done to get a network that (i) accumulates temporal information internally for proper rate estimation, and (ii) can keep estimating stably beyond the length of the window it was trained on. We train until convergence using an MSE (mean squared error) loss, Adam optimizer and a learning rate of 1e-4. We add a weight of 10 to the attitude loss to make it similar in magnitude to the rate loss. Data augmentation consists of taking random slices of frames, randomly flipping event frames (and labels accordingly) in the channel dimension (polarity flips), height (up-down flips) and width (left-right flips). For evaluation, we run networks on full sequences without resetting, and we evaluate in terms of RMSE (root mean square error) and MASD (mean absolute successive difference):
\begin{align}
    \text{RMSE} &= \sqrt{\frac{1}{n}\sum_{i=1}^n \left(y_i - \hat{y}_i\right)^2} \\
    \text{MASD} &= \frac{1}{n-1}\sum_{i=1}^{n-1} \lvert x_{i+1} - x_i \rvert
\end{align}

The baseline network has 425k parameters and consists of an 8$\times$-downsampling encoder followed by a GRU (gated recurrent unit) memory block and a decoder transforming the memory into angle and rate predictions. The encoder and memory are convolutional to stimulate the learning of local motion-related features that generalize well, and to prevent overfitting to scene-specific appearance (which we observed when using a fully connected GRU). We experiment with different network variants in terms of receiving extra inputs (drone motor speeds, rotation rates) or having a feedforward instead of recurrent block (no memory). Extra inputs give a slightly larger network with 453k paramters, while the removal of recurrent connections gives a network of only 240k parameters. Schematic illustrations of these are shown in \cref{fig:networks}. Apart from the GRU blocks and final output, we use ELU (exponential linear unit)~\cite{clevert2016fast} activations throughout the network.

\subsection*{Using estimates for real-world robot control}

We use the two-layered control architecture illustrated in \cref{fig:overview}. While it closely follows the implementation in our flight controller, it is running on a separate on-board computer. This allows us to use the control gains from the flight controller as-is, with only minor tweaking to account for the delay added by communication between the on-board computer and the flight controller.

The first layer consists of a proportional controller that compares commanded attitude and estimated attitude to generate rotation rate setpoints. The second layer is implemented as a PID (proportional-integral-derivative) controller, translating these rotation rate setpoints---combined with estimated rotation rates---into torque commands. These are then sent to the motor mixer running on the flight controller. The motor mixer converts these torque commands, along with a single thrust command, into individual motor commands. The mixing process involves a linear transformation based on the specific geometric layout of the drone.

For our flight tests, we have a human pilot controlling the drone. While stable flight is possible with the pilot immediately commanding the attitude of the drone (angle or stabilized mode), we make use of an outer-loop position controller running on the flight controller (position mode). An additional optical flow sensor provides velocity estimates, allowing the pilot to control position instead of attitude. This greatly simplifies flight testing, and makes individual tests more repeatable. Furthermore, because our training data inherently contains biases, some kind of outer-loop controller (whether a human pilot or a software position controller) is necessary to mitigate these biases during testing.

\subsubsection*{Robot setup}

We use a custom 5-inch quadrotor (shown in \cref{fig:overview}) to perform real-world flight tests. The drone has a total weight of approximately 800~g, including sensors, actuators, on-board compute and battery. All algorithms are implemented to run entirely on board, using an NVIDIA Jetson Orin NX embedded GPU to receive data from the event camera, estimate flight attitude and rotation rate, and calculate control commands in real time. Body torque commands based on the estimated attitude and rotation rate are sent to the flight controller, which is a Kakute H7 mini running the open-source autopilot software PX4. Communication between the flight controller and the embedded GPU is done using ROS2~\cite{macenski2022robota}. An MTF-01 optical flow sensor and rangefinder enables pilot control in position mode. We record ground-truth flight trajectories (for plotting) using a motion capture system.

We use a downward-looking DVXplorer Micro event camera in combination with a 140\textdegree-field-of-view lens to capture as much of the environment as possible. To prevent bandwidth saturation while keeping good estimation performance, we only enable every other pixel on the sensor, resulting in a 320$\times$240 stream (instead of 640$\times$480) for the same field-of-view. These events are accumulated into 5~ms frames for the network. The entire events-to-attitude pipeline is running at approximately 200~Hz, with the embedded GPU consuming around 9~W on average. We found that at least 200~Hz is necessary for stable control, due to the torque commands going directly to the motors. While the pipeline itself can run at frequencies as high as 1000~Hz, communication limited real-world flight tests to 500~Hz without any logging, and 200~Hz with logging.

\backmatter

\bmhead{Data and code availability} 

Datasets, code and hardware setup instructions are publicly available (upon publication) via the project's webpage: \url{https://mavlab.tudelft.nl/attitude_from_events}.

\bmhead{Acknowledgements}

We would like to thank the participants of the 2024 Capo Caccia Neuromorphic Workshop for their discussions and insights. This work was supported by funding from the Air Force Office of Scientific Research (award no. FA8655-20-1-7044) and the Dutch Research Council (NWO, NWA.1292.19.298).

\bmhead{Author contributions}

All authors contributed to the conception of the project and to the analysis and interpretation of the results. J.H. and S.S. built the drone and integrated hardware and software into a stably flying platform. J.H. developed the software for training networks and running them on board the drone. S.S. developed the software for converting network estimates to low-level control commands on the drone. J.H. and S.S. collected datasets for training and performed flight tests to gather results. J.H. and S.S. wrote the first draft of the article. All authors contributed to the reviewing of draft versions and gave final approval for publication.

\bibliography{sn-bibliography}%

\end{document}